**RESEARCH**

**Open Access**

# Large-scale protein-protein post-translational modification extraction with distant supervision and confidence calibrated BioBERT

Aparna Elangovan[1], Yuan Li[1], Douglas E. V. Pires[1], Melissa J. Davis[2,3] and Karin Verspoor[1,4*]

*Correspondence:
karin.verspoor@rmit.edu.au
[4] School of Computing Technologies, RMIT University, Melbourne, Australia
Full list of author information is available at the end of the article

## Abstract

**Motivation:** Protein-protein interactions (PPIs) are critical to normal cellular function and are related to many disease pathways. A range of protein functions are mediated and regulated by protein interactions through post-translational modifications (PTM). However, only 4% of PPIs are annotated with PTMs in biological knowledge databases such as IntAct, mainly performed through manual curation, which is neither time- nor cost-effective. Here we aim to facilitate annotation by extracting PPIs along with their pairwise PTM from the literature by using distantly supervised training data using deep learning to aid human curation.

**Method:** We use the IntAct PPI database to create a distant supervised dataset annotated with interacting protein pairs, their corresponding PTM type, and associated abstracts from the PubMed database. We train an ensemble of BioBERT models—dubbed PPI-BioBERT-x10—to improve confidence calibration. We extend the use of ensemble average confidence approach with confidence variation to counteract the effects of class imbalance to extract high confidence predictions.

**Results and conclusion:** The PPI-BioBERT-x10 model evaluated on the test set resulted in a modest F1-micro 41.3 (P = 5 8.1, R = 32.1). However, by combining high confidence and low variation to identify high quality predictions, tuning the predictions for precision, we retained 19% of the test predictions with 100% precision. We evaluated PPI-BioBERT-x10 on 18 million PubMed abstracts and extracted 1.6 million (546507 unique PTM-PPI triplets) PTM-PPI predictions, and filter ≈ 5700 (4584 unique) high confidence predictions. Of the 5700, human evaluation on a small randomly sampled subset shows that the precision drops to 33.7% despite confidence calibration and highlights the challenges of generalisability beyond the test set even with confidence calibration. We circumvent the problem by only including predictions associated with multiple papers, improving the precision to 58.8%. In this work, we highlight the benefits and challenges of deep learning-based text mining in practice, and the need for increased emphasis on confidence calibration to facilitate human curation efforts.

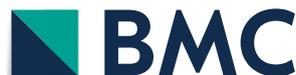

## Background

Critical biological processes, such as signaling cascades and metabolism, are regulated by protein-protein interactions (PPIs) that modify other proteins in order to modulate their stability or activity via post-translational modifications (PTMs). PPIs are curated in large online repositories such as IntAct [1] and HPRD [2]. However, most PPIs are not annotated with a function, for example, we found the IntAct database has over 100,000 human PPIs, but less than 4000 of these are annotated with PTMs such as phosphorylation, acetylation or methylation. Understanding the nature of PTM between an interacting protein pair is critical for researchers to determine the impact of network perturbations and downstream biological consequences. PPIs and PTMs in biological databases are usually manually curated, which is time consuming and requires highly trained curators. [3]. Orchard et al. [4] have also highlighted additional challenges in maintaining manually curated databases, ensuring they are up to date, as well as the economic aspects of manual curation. Hence, the adoption of automated curation methods is essential for sustainability of this work.

Here we extract PTMs by text mining PubMed abstracts, extracting protein pairs along with their corresponding PTM. Given an input journal abstract, the output is a triplet of the form <Protein1, PTM function, Protein2> where Protein1 and Protein2 are Uniprot identifiers [5] of the proteins. Since our training data source does not contain the direction of the relationship between the proteins, we do not take into account the direction of relationship between Protein1 and Protein2, i.e.<Protein1, PTM function, Protein2> is equivalent to <Protein2, PTM function, Protein1>. We also aim to aid human curation of PTM-PPIs, hence we assess how well machine learning models generalise by applying them to 18 million PubMed abstracts to extract PTM-PPI triplets. In this paper, we use confidence calibration [6, 7] as a mechanism to understand generalisability to know when a prediction works to extract high quality predictions. We believe our paper is the first to study the practical applicability and challenges of large scale PTM-PPI extraction using NLP with deep learning and distant supervision.

We focus on extracting PTMs including phosphorylation, dephosphorylation, methylation, ubiquitination, deubiquitination, and acetylation ( these PTMs were selected based on the availability of training data). We create a training dataset using distant supervised approach [8, 9] using IntAct [1] as the source knowledge base to extract PTM-PPI triplets from PubMed abstracts. We train an ensemble of BioBERT [10] models to improve neural confidence calibration [6, 7]. We then applied the trained model on 18 million PubMed abstracts to extract PPI pairs along with their corresponding PTM function and attempt to ensure high quality predictions using neural confidence calibration techniques [6, 7] to augment and facilitate human curation efforts.

### Related works in protein interaction extraction through deep learning

The datasets in PPI extraction such as AIMed [11] and BioInfer [12] used to evaluate text mining approaches have remained the same for over a decade (since 2007) and focus on extracting protein interactions but not the nature of the PTM interaction between them. These datasets have also been used to evaluate the latest machine learning approaches including deep learning [13, 14] in protein pair extraction. However, the latest deep learning trends do not seem to be widely popular in PPI curation outside the limited context of benchmarking methods using the AIMed [11] and BioInfer [12] datasets. Automated PPI curation attempts using text mining and rule-based approaches seem more prevalent [15–17].

STRING v11 [18], one of the most popular PPI databases, uses text mining as a curation method. Their text mining pipeline has largely remained the same since STRING v9.1 [15]. STRING v9.1 uses a weighted PPI co-occurrence rule-based approach, where the weights depend on whether a protein pair occurs together within the same document, the same paragraph and/or the same sentence. Rule based approaches can be quite effective [15, 19] even with limited training data, depending on the task. For instance, Szklarczyk et al. define an interaction unit in STRING v11 database [18] as a "functional association, i.e. a link between two proteins that both contribute jointly to a specific biological function". This definition allows for co-occurrence rule-based approach to be quite effective, i.e. if a protein pair co-occurs in text frequently then the pair is highly likely to be related.

iPTMnet [20] consolidates information about PPIs and PTMs from various manually curated databases such as HPRD [2] and PhosphoSitePlus [21] as well as text mining sources. For text mining, iPTMnet uses RLIMS-P [17] and eFIP [16] to automatically curate enzyme-substrate-site relationships. These tools use rule-based approaches using text patterns to extract proteins involved in PTM. The iPTMnet statistics dated Nov 2019 (https://research.bioinformatics.udel.edu/iptmnet/stat) indicate that the total number of enzyme-substrate pairs curated using RLIMS-P [17] is fewer than 1,000 pairs. This modest number highlights the main challenge using text patterns: while they can extract relationships with fairly high precision, they are not robust to variations in how PPI relations can be described in text. We therefore explore machine learning-based methods, which have been able to extract substantially more relationships as seen in PTM site extraction work by Björne et al. [22].

Automatic extraction of PPIs using deep learning can be beneficial as it has the potential to extract PPIs from a variety of text where the PPI relationships are described in ways that cannot be easily captured by a manually crafted rule-based system. However, deep learning requires a much larger volume of training data. Generalisability of the model to ensure prediction quality is key to its widespread adoption for automatic extraction of PPI relationships from text at scale. Enhancing quality of prediction at large scale needs to focus on reducing false positives to minimise corrupting existing knowledge base entries (i.e. predicting that a relationship exists when it doesn't) and, hence, confidence calibration approaches to reduce poor quality predictions become a crucial step in large scale text mining. Confidence calibration is the problem of predicting probability estimates representative of the true correctness [7] and in this paper we use confidence calibration to know when a prediction is likely correct and use it as a mechanism to improve generalisation. The aspects of generalisability have been largely limited to the evaluation of a test set and the limitations of using the test set performance as a proxy for real world performance have been challenged in previous studies [23, 24].

Creating gold standard training data with fine-grained annotations is a manual, labor-intensive task and is a limiting factor in applying machine learning to new domains or tasks. Being able to leverage one or more existing data sources is key to using machine learning in new domains or for new tasks. Distant supervision [8, 9] exploits existing knowledge bases, such as IntAct [1], instead of annotating a new dataset. However there are two main limitations to the use of distant supervision datasets: (a) noisy labels require noise reduction techniques to improve label quality [25] (b) they require negative samples to be generated as the databases usually only contain positive examples of a relationship.

Deep learning architectures such as BiLSTM and BioBERT have been previously used to benchmark methods for protein relation extraction using Natural Language Processing (NLP) and the AIMed dataset [13, 23, 26]. However, these works [13, 23, 26] do not measure the ability of these models to calibrate confidence scores. We chose a state-of-the-art deep learning approach, BioBERT [10], train an ensemble to enhance confidence calibration [6] and use confidence variation to counteract the effects of class imbalance during confidence calibration.

## Methods

### Dataset

We obtain the dataset of human PTM-PPI interactions from the IntAct database [1], a database that is part of the International Molecular Exchange (IMEX) [27] Consortium. The database contains Uniprot identifiers, protein aliases, interaction type (where available), and the PubMed identifier of the paper describing the interaction. Of the over 100,000 human PPIs in IntAct [1], only 3381 PPIs describe PTM interactions. Hence, we start with an initial dataset containing these 3381 PPIs. We then remove duplicate entries, i.e.the interacting proteins, the PubMedId and the interaction type are the same resulting in 2797 samples. We then pre-process this data to remove self-relations (where participant protein1 is the same as protein2), as they tend to degrade performance [19].

Once we pre-process the data, the steps involved in dataset transformation is illustrated in Table 1. We first identify all the gene mentions in an abstract and their corresponding Uniprot protein identifier [5], as detailed in section *Gene mentions and protein identifiers*. We apply noise reduction techniques to remove samples where the abstract may not describe the PPI relationship, as detailed in section *Noise reduction*. We then split the dataset into train, test and validation sets such that they are stratified by interaction type and have unique PubMed ids in each set to avoid test set leakage resulting from random splits identified in the AIMed dataset [23]. Negative training samples are generated as detailed in section *Negative sample generation*.

### Gene mentions and protein identifiers

To prepare the training data, the Uniprot identifiers mentioned in IntAct need to match the Uniprot identifiers that can be inferred from the abstract text by applying named entity recognition and normalisation NLP techniques, *gene mention* → *Uniprot accession number*. GNormPlus [28] identifies gene names and normalises them to the NCBI gene identifier [29]. The NCBI gene identifier is a gene-level identifier that needs to be converted to a Uniprot accession code which is a protein identifier. A single NCBI gene can be associated with multiple Uniprot accession numbers either due to biological reasons such as alternative splicing where a single gene results in different protein isoforms or due to technical reasons related to Uniprot accession number changes [5]. For a given

**Table 1** Illustration of data preparation

| | |
|---|---|
| **IntAct** | **Pubmed Id: 24291004 , Uniprot protein pair: P04150, *P31749*, Interaction type: Phosphorylation** |
| Corresponding abstract | *Glucocorticoid resistance ..[truncated display]....* we identify the AKT1 kinase as a major negative regulator of the NR3C1 glucocorticoid receptor protein activity driving glucocorticoid resistance in T-ALL. Mechanistically, **AKT1 impairs glucocorticoid-induced gene expression by direct phosphorylation of NR3C1** at position S134 and blocking glucocorticoid-induced NR3C1 translocation to the nucleus. Moreover, we demonstrate that loss of PTEN and consequent AKT1 activation can effectively block glucocorticoid-induced apoptosis and induce resistance to glucocorticoid therapy. Conversely, pharmacologic inhibition of AKT with MK2206 effectively restores glucocorticoid-induced NR3C1 translocation .... *in vitro and in vivo*. |
| Genes and normalised NCBI ids | • **Start-End, Gene Mention, NCBI gene Id**<br>• 137 - 141, AKT1, 207<br>• 186 - 191, NR3C1, 2908<br>• ***294 - 298, AKT1, 207***<br>• ***375 - 380, NR3C1, 2908***<br>• 434 - 439, NR3C1, 2908<br>• 508 - 512, PTEN, 5728<br>• 528 - 532, AKT1, 207<br>• 748 - 753, NR3C1, 2908 |
| NCBI to Uniprot map | • **NCBI gene: Uniprot identifiers**<br>• **2908**: E5KQF5, E5KQF6, F1D8N4, **P04150**, B7Z7I2<br>• *207*: B0LPE5, **P31749**, B3KVH4<br>• *5728*: P60484, F6KD01 |
| Normalised abstract for pair **P04150**, *P31749* | *Glucocorticoid resistance ....[truncated display]....* we identify the **P31749** kinase as a major negative regulator of the **P04150** glucocorticoid receptor protein activity driving glucocorticoid resistance in T-ALL. Mechanistically, *P31749* impairs glucocorticoid-induced gene expression by direct phosphorylation of **P04150** at position S134 and blocking glucocorticoid-induced **P04150** translocation to the nucleus. Moreover, we demonstrate that loss of *P60484* and consequent **P31749** activation can effectively block glucocorticoid-induced apoptosis and induce resistance to glucocorticoid therapy. Conversely, pharmacologic inhibition of AKT with MK2206 effectively restores glucocorticoid-induced **P04150** translocation ..... *in vitro and in vivo*. |
| Negative sample pairs | • **P04150**, *P60484*<br>• *P31749*, *P60484* |

The bold highlights where the relationship is specified within the abstract

The IntAct database has the PubMedId, the Uniprot identifiers of the participating proteins and the interaction type. More than 1 pair of interacting proteins can be annotated against a given PubMed Id and not all of these interactions are described in the abstract. This forms the noisily labelled training data abstract $A$, and Uniprot identifiers of the two PPI participants $IU_1, IU_2$ annotated in the IntAct database as $< A, IU_1, IU_2 >$, we perform the following:

1. Use GNormPlus to identify gene name mentions $[g_1, g_2..g_m]$ in abstract $A$, and to normalise gene names to corresponding NCBI gene ids, $NG = [ng_1, ng_2, \ldots, ng_m]$.
2. Given a NCBI gene id, $ng_i \in NG$, we obtain a list of corresponding Uniprot accession numbers $U = [u_1, u_2, \ldots u_n]$.
3. If $IU_1$ or $IU_2$ exists in $U$, then we have found a matching IntAct Uniprot identifier $IU_x$ annotated against the abstract, where $IU_x \in [IU_1, IU_2]$. Map NCBI gene id $ng_i$ to Uniprot identifier $IU_x$. If neither $IU_1$ nor $IU_2$ exist in $U$, i.e.the Uniprot identified in the abstract is not annotated against the abstract, then simply return the first Uniprot identifier $u_1$.

This gene mention identification to Uniprot identifier mapping process is illustrated through an example in Table 1.

#### Noise reduction

In order to extract PTM-PPI relationships mentioned in the abstract, we aim to retain only those training samples where the PTM-PPI is described in the abstract. Hence we exclude IntAct PTM-PPI triplets from the training data if either participant's Uniprot identifiers does not exist in the normalized abstract. This is to minimize the false positive noise in the distant supervised dataset, i.e.removing samples where the relationship is not described in the abstract, based on the assumption that, if the proteins are not explicitly mentioned in the abstract then it is highly likely (unless protein names are not recognized by the NER tool) that the abstract does not describe the relationship between those proteins. We also remove PPIs where the stemmed interaction type (phosphoryl, dephosphoryl, methyl, acetyl, deubiquitin, ubiquitin, demethyl) is not mentioned in the abstract.

#### Negative sample generation

Since knowledge bases focus on relationships that exist, distant supervision-based approaches require negative training examples to be created. In our dataset, negative samples are protein pairs that are mentioned in the abstract but do not have a function referencing that paper annotated against the pair in Intact. In order to generate negative samples, we identify protein mentions from the abstracts using GNormPlus [28], which normalizes mentions to NCBI gene IDs which are then converted to Uniprot identifiers as described in section *Gene mentions and protein identifiers*. If a given protein pair $<p1, p2>$, where $p1$ and $p2$ are mentioned in the abstract, but is not annotated against any of the 7 types of PPI relationship within the abstract according to the Intact database, then it is assumed that $<p1, p2>$ form negative samples for that abstract, see example in Table 1.

It is important to emphasize that a negative sample does not mean that a given PPI relationship does not exist, but rather that the abstract does not describe such a relationship. It could also be a noisy negative sample, i.e.the abstract describes the functional relationship between pair, but it is simply not captured in the annotation.

#### Training BioBERT for PPI extraction

We fine-tuned the pretrained BioBERT v1.1 [10] is pretrained on a large collection of PubMed abstracts. We applied fine-tuning to adapt BioBERT to the PTM-PPI extraction task, as a multi-class classification problem, and assumed there is at most one type of PTM relationship between a protein pair. The multi-class task has 7 labels, 6 PTMs and 1 negative label. Since the training data is imbalanced, as the training progresses through epoch the best model is snapshot-ed based on the highest F1-macro, instead of lowest cross entropy loss. For more details on BioBERT training configuration, see Additional file 1: Section A.1.

Before feeding the input to the model, the input normalised abstract is transformed such that the UniProt identifiers of the participating proteins are replaced with fixed marker names `<PROTPART1, PROTPART2>`, and the names of all other proteins were replaced with `<PRTIG1, PRTIG2,...PRTIGn>` where *n* is the total number of unique non-participating UniProt identifiers within the normalised abstract. For instance, given a training sample i.e.the normalised abstract [*..P31749 impairs* *glucocorticoid-induced gene expression by direct phosphorylation of P04150 and blocking glucocorticoid-induced P04150 translocation ... Moreover, we demonstrate that loss of P60484.... and P60484...*] and triplet `<P31749, phosphorylation, P04150>` the transformation would result in [..`PROTPART1` *impairs glucocorticoid-induced gene expression by direct phosphorylation of* `PROTPART2` *and blocking glucocorticoid-induced* `PROTPART2` *translocation ... Moreover, we demonstrate that loss of PRTIG1 .... and PRTIG1...*]. This approach is similar to previous pre-processing techniques used in PPI extraction [13, 23].

The original BioBERT is trained using a sentence as a training sample for the language modelling task, while we work with complete abstracts. To utilize an entire abstract as a single training sample, we feed all the sentences within the abstract including the full stop (end-of-sentence marker, the period ".") separating each sentence. BioBERT has a maximum input sequence length of 512 units, this includes the sub word units to fit the entire abstract. Sub word units are concatenated to represent words and limit the vocabulary size, *e.g.*, a word such as Immunoglobulin is tokenized into 7 units (I, ##mm, ##uno, ##g, ##lo, ##bul, ##in) where ## are special symbols to indicate continuation [30]. Therefore, an abstract with 250 words can result in a much longer sequence once it is tokenised. In our training data, post tokenisation, 90% of the normalised abstracts are under the maximum limit of 512 sub units. For more detailed distribution see Additional file 1: Table S10. Therefore, we simply truncate sequences longer than 510, and reserve 2 units to accommodate the mandatory marker [CLS] and [SEP] tokens that BERT requires [30], where the [SEP] token is used to separate two parts of the input, and the [CLS] token is used to capture the overall representation of the input.

**Uncertainty estimation**

In order to improve the probability estimate associated with each prediction, we use an ensemble of 10 PPI-BioBERT models (referred to as PPI-BioBERT-x10), all trained with exactly the same hyperparameters and training data but capturing slightly different models due to Bernoulli (binary) dropout [31] layers in the network. The use of ensembles of models to improve uncertainty estimates, rather than just improving overall accuracy, has been shown to be effective in the computer vision task of image classification by Lakshminarayanan et al. [6], hence we follow that approach. Using the ensemble of the 10 models, the predicted confidence $\hat{p}_j$ and predicted class $\hat{y}_j$ for the input $x_j$ is:

$$p(y_j = c) = \frac{1}{M} \Sigma_{i=1}^{M} p(y_{\theta_i} = c | x_j, \theta_i) \quad (1)$$

$$\hat{p}_j = max_c p(y_j = c) \quad (2)$$

$$\hat{y}_j = argmax_c p(y_j = c) \quad (3)$$

In addition to averaging confidence as proposed by Lakshminarayanan et al. [6], we also use the standard deviation of the predicted confidence across the ensemble as:

$$std(\hat{p}_j) = \sqrt{\frac{1}{M}\Sigma_{i=1}^{M}(p(y_{\theta_i} = \hat{y}_j|x_j, \theta_i) - \hat{p}_j)^2} \quad (4)$$

where M is the number of models in the ensemble; $\theta_i$ is model i; $y_{\theta_i}$ is the output predicted by model i; x is the input; c is the predicted class. Lakshminarayanan et al. [6] also use adversarial samples in their image classification tasks. We do not use adversarial samples in our experiment as Lakshminarayanan et al. [6] show that adversarial examples make very little difference to accuracy as the ensemble size increases beyond 5. In addition, effective adversarial examples are harder to create for NLP tasks compared to image classification.

We measure the ability of the ensemble in confidence calibration using Expected Calibration Error (ECE) used previously by Guo et.al. [7] in computer vision. ECE is defined as:

$$accu(B_k) = \frac{1}{|B_k|}\Sigma_{j \in B_k} 1(\hat{y}_j = y_j) \quad (5)$$

$$conf(B_k) = \frac{1}{|B_k|}\Sigma_{j \in B_k} \hat{p}_j \quad (6)$$

$$ECE = \Sigma_{k=1}^{K} \frac{|B_k|}{n}|accu(B_k) - conf(B_k)| \quad (7)$$

where the confidence score is divided into K equally spaced bins $B_k$, n is the total number of samples (intuitively $\frac{|B_k|}{n}$ represents the fraction of samples that fall into the bin $B_k$), $accu(B_k)$ is the accuracy of predictions whose confidence fall into bin $B_k$ and $conf(B_k)$ is the average confidence of predictions that confidence fall into bin $B_k$. The best calibrated model will have zero ECE.

**Large scale extraction from PubMed abstracts**

We extracted the complete collection of 18 million abstracts available in PubMed as of April 2019. We applied GNormPlus [28] to recognize the proteins and normalize them to UniProt identifiers. We then apply the PPI-BioBERT-x10 model to extract PTM-PPI relationships from the entire PubMed abstracts. For inference, given a normalised abstract with *n* unique UniProt identifiers mentioned in the abstract we generate $C(n, 2) = \frac{n!}{(n-2)!2!}$ transformed inputs, where the inputs are transformed as detailed in section *Training BioBERT for PPI extraction*, to predict the relationship between every possible combination of protein pair except self-relations. In order to extract high quality PTM-PPI relationships, we only select predictions that have low variation in the confidence score and high prediction confidence.

## Results

### IntAct-based dataset

The final dataset includes the Uniprot identifiers of the participants, the normalized Abstract (the protein names normalised to Uniprot identifiers), all the proteins identified through NER in the abstract and the class label (the type of PTM interaction). After

**Table 2** Train/test/val positive samples for each interaction type.

|  | Train | | Val | | Test | | Total | |
| --- | --- | --- | --- | --- | --- | --- | --- | --- |
|  | − | + | − | + | − | + | − | + |
| Acetylation | 31 | 5 | 2 | 1 | 9 | 1 | 42 | 7 |
| Dephosphorylation | 167 | 28 | 36 | 10 | 33 | 6 | 236 | 44 |
| Deubiquitination | 4 | 2 | 0 | 0 | 0 | 0 | 4 | 2 |
| Methylation | 22 | 10 | 5 | 1 | 22 | 4 | 49 | 15 |
| Phosphorylation | 862 | 139 | 118 | 21 | 227 | 44 | 1207 | 204 |
| Ubiquitination | 30 | 5 | 5 | 1 | 5 | 1 | 40 | 7 |
| Total | 1116 | 189 | 166 | 34 | 296 | 56 | 1578 | 279 |

The negative samples count against a interaction type is solely to indicate how many were derived from abstracts describing the interaction type. All negative samples belong to a single class *"other/Negative"* pre-processing and applying noise reduction to the original dataset with approximately over 3000 interactions in IntAct, we are left with a total of 279 positive and 1579 negative samples across training, test and validation set. The distribution of interaction types and positive/negative samples is shown in Table 2, with phosphorylation forming almost 75% of the positive samples in the training set. The positive sample rate is between 14–17% in the train, test and validation set. Interaction types such as such acetylation, deubiquitination and ubiquitination, have less than 10 positive samples in total across training, test and validation.

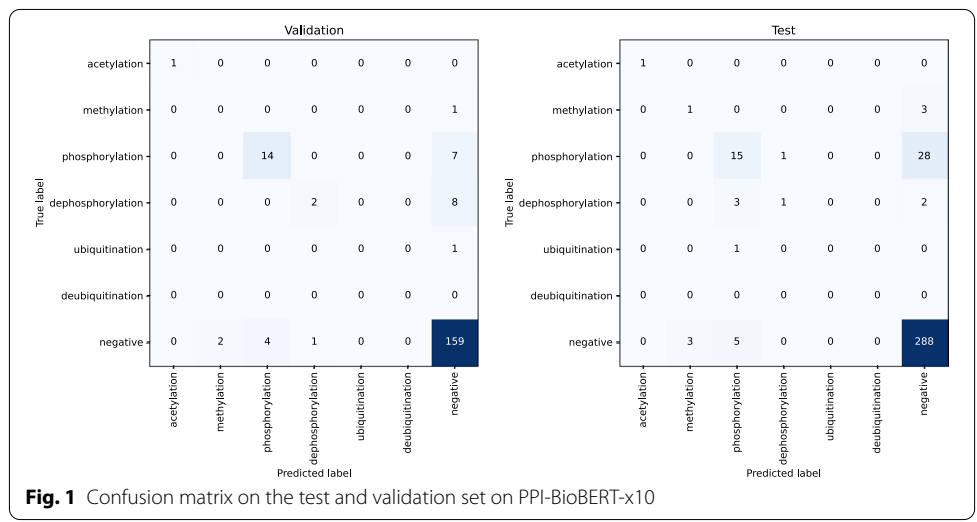

**Fig. 1** Confusion matrix on the test and validation set on PPI-BioBERT-x10

**Confidence calibration on the test and validation sets**

The performance on the test and validation sets using the ensemble BioBERT model (PPI-BioBERT-x10) is detailed in Table 3. The test and validation sets have an F1-micro score of 41.3 and 58.6, respectively, across all interactions. The confusion matrix is shown in Fig. 1.

We visualise the ensemble model's confidence calibration using reliability diagrams, similar to Guo et al. [7], as shown in Fig. 2. We observe that the predicted confidence range for each interaction type is proportional to the percentage of training samples, i.e., the negative samples have the highest proportion and over 80% of the negative sample

**Table 3** The performance of ensemble PPI-BioBERT-x10 on the test and validation set

| Dataset | Interaction | P | R | F1 | ECE | SD | Support |
|---|---|---|---|---|---|---|---|
| Test | Acetylation | 100.00 | 100.00 | 100.00 | 0.49 | 0.25 | 1 |
| Test | Dephosphorylation | 50.00 | 16.67 | 25.00 | 0.67 | 0.40 | 6 |
| Test | Methylation | 25.00 | 25.00 | 25.00 | 0.60 | 0.28 | 4 |
| Test | Phosphorylation | 62.50 | 34.09 | 44.12 | 0.79 | 0.26 | 44 |
| Test | Ubiquitination | 0.00 | 0.00 | 0.00 | – | – | 1 |
| Test | ECE | – | – | - | 0.75 | – | 31 |
| Test | Average SD | – | – | – | – | 0.28 | 31 |
| Test | Macro avg | 47.50 | 35.15 | 38.82 | – | – | 56 |
| Test | Micro avg | 58.06 | 32.14 | 41.38 | – | - | 56 |
| Val | Acetylation | 100.00 | 100.00 | 100.00 | 0.53 | 0.16 | 1 |
| Val | Dephosphorylation | 66.67 | 20.00 | 30.77 | 0.61 | 0.37 | 10 |
| Val | Methylation | 0.00 | 0.00 | 0.00 | 0.53 | 0.29 | 1 |
| Val | Phosphorylation | 77.78 | 66.67 | 71.79 | 0.78 | 0.26 | 21 |
| Val | Ubiquitination | 0.00 | 0.00 | 0.00 | – | – | 1 |
| Val | ECE | – | – | – | 0.73 | – | 24 |
| Val | Average SD | – | – | – | – | 0.28 | 24 |
| Val | Macro avg | 48.89 | 37.33 | 40.51 | – | – | 34 |
| Val | Micro avg | 70.83 | 50.00 | 58.62 | – | – | 34 |

**ECE** is the expected calibration error. **SD** denotes the average standard deviation within the ensemble

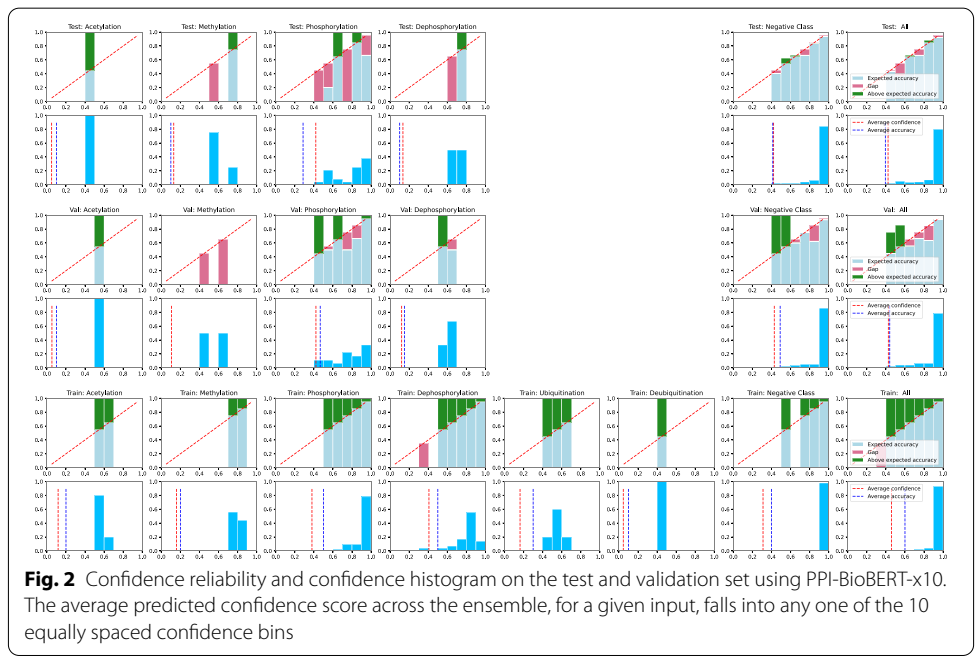

**Fig. 2** Confidence reliability and confidence histogram on the test and validation set using PPI-BioBERT-x10. The average predicted confidence score across the ensemble, for a given input, falls into any one of the 10 equally spaced confidence bins

predictions have a confidence score between 0.9 and 1.0, whereas acetylation, which has less than 1% of the training samples, has prediction confidence of less than 0.6. This relationship between the distribution of predicted confidence and the proportion of training samples for a given class label (interaction type) is consistent in the train, validation and test set as shown in Fig. 2. This highlights 2 main limitations of measuring calibration

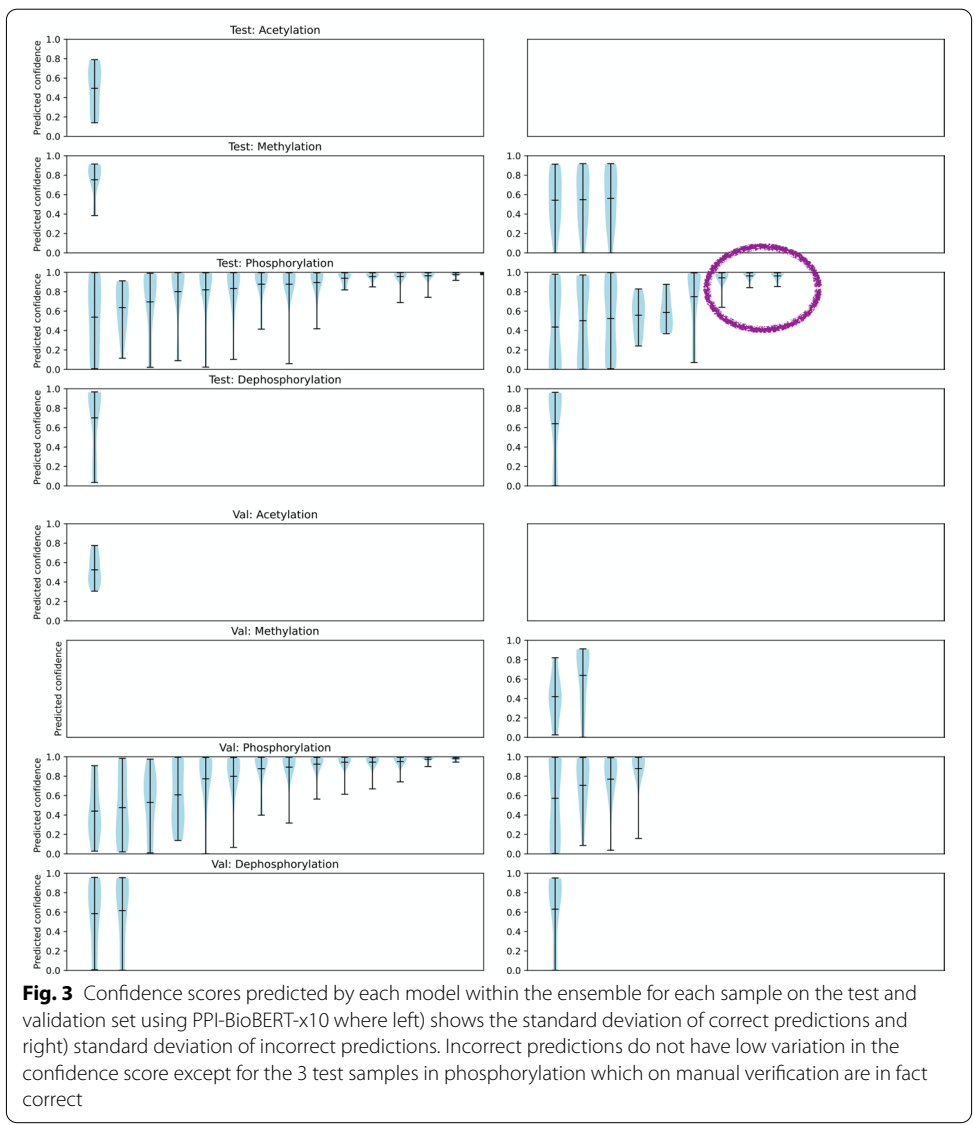

**Fig. 3** Confidence scores predicted by each model within the ensemble for each sample on the test and validation set using PPI-BioBERT-x10 where left) shows the standard deviation of correct predictions and right) standard deviation of incorrect predictions. Incorrect predictions do not have low variation in the confidence score except for the 3 test samples in phosphorylation which on manual verification are in fact correct

error using ECE: **(a)** it penalises model calibration regardless of the class imbalance in the training dataset when high quality predictions can be available at lower confidence scores for classes with lower proportion of samples **(b)** it penalises both over- and under-calibration equally. Hence, we find it challenging to rely on ECE and/or average ensemble confidence to extract high quality predictions, especially for interaction types that have a much lower proportion of training samples.

We therefore inspected the variation in the confidence score predicted by each model for a given sample as shown in Fig. 3. Intuitively, if for a given input, all the models within the ensemble consistently predict with similar confidence then we can rely on the ensemble confidence better compared to high variation confidence scores. When we compare the variation in confidence scores of the correct versus incorrect predictions, the incorrect predictions do not seem to have low variation in predicted confidence, as shown in Fig. 3, except for 3 incorrect phosphorylation predictions. We apply a heuristic

**Table 4** Manually verified relationships in the validation and test set with ensemble prediction standard deviation less than interaction-wise threshold and predicted confidence greater than interaction-wise threshold, taking the precision to 100.0

| PubMed | Phrases describing relationships in the abstract | Label | Prediction |
|---|---|---|---|
| Validation | | | |
| 12150926 | **mTOR (P42345)**-catalyzed phosphorylation of **4EBP1 (Q13541)** in vitro | phosphorylation | phosphorylation |
| 15733869 | **SGK (O00141)** physically associates with **CREB (P16220)** and **SGK (O00141)** phosphorylates it on serine 133 | phosphorylation | phosphorylation |
| 15557335 | **Src (P12931)** phosphorylated **Alix (Q8WUM4)** at a C-terminal | phosphorylation | phosphorylation |
| 15527798 | Phosphorylation of **MDM2 (Q00987)** by the protein kinase **AKT (P31749)** | phosphorylation | phosphorylation |
| 10864201 | Radiation-induced phosphorylation of **p53 (P04637)** protein at serine 15, largely mediated by **ATM (Q13315)** kinase | Phosphorylation | phosphorylation |
| 24548923 | **Akt (P31749)**-mediated phosphorylation of Carma1 **(Q9BXL7)** | phosphorylation | phosphorylation |
| 19407811 | **BubR1 (O60566)** forms a complex with **PCAF (Q92831)** and is acetylated at lysine 250 | acetylation | acetylation |
| Test | | | |
| 21920476 | **coilin (P38432)** is phosphorylated in Ser184 by both VRK1 (Q99986) | phosphorylation | phosphorylation |
| 19424295 | Previous studies of **cofilin (P23528)** have shown that it is phosphorylated primarily by the LIM domain kinases **Limk1 (P53667)** | phosphorylation | phosphorylation |
| 22726438 | **FGFR2 (P21802)** phosphorylates tyrosine residues on **Grb2 (P62993)** | phosphorylation | phosphorylation |
| 11154276 | **Akt (P31749)** decreased **ASK1 (Q99683)** kinase activity stimulated by both oxidative stress and overexpression in 293 cells by phosphorylating a consensus **Akt (P31749)** site at serine 83 of **ASK1 (Q99683)** | phosphorylation | phosphorylation |
| 25605758 | Phosphorylation of **Rab5b (P61020)** by LRRK2 **(Q5S007)** also exhibited | phosphorylation | phosphorylation |
| 20856200 | Binding of **AKT (P31749)** (tail region) to **Vim (P08670)** (head region) results in **Vim (P08670)** Ser39 phosphorylation | phosphorylation | phosphorylation |
| 21986944 | **MAK (P20794)** associates with **CDH1 (P12830)** (FZR1 (Q9UM11), fizzy/cell division cycle 20 related 1) and phosphorylates **CDH1 (P12830)** | Negative | phosphorylation |
| 15862297 | **HDM2 (Q00987)** phosphorylation by Chk2 (O96017) doubles in the presence of **p53 (P04637)** | Negative | phosphorylation |
| 21887822 | **Hsp70 (P34932)** is phosphorylated by **Plk1 (P53350)** | Negative | phosphorylation |

The bold highlights where the mentions of participating proteins to select interaction-wise confidence and standard deviation thresholds where we select the 50th percentile as the cut-off for both confidence and confidence standard deviation based on the training set. This results in retaining 19% of positive predictions (6 out of 31) in the test set. Since our training data is noisy, we manually verify all predictions that have low variation in the test and validation set. On manual verification, we find all the predictions using this heuristic are in fact correct, including 3 incorrectly labelled ones, as shown in Table 4, indicating that a combination of low variation and relative high confidence has the potential to be robust against noisy labels.

**Large scale PTM-PPI extraction from PubMed abstracts**

We extracted approximately 1.6 million PPIs (546507 unique PTM-PPI triplets) from 18 million abstracts using the PPI-BioBERT-x10 model, see Table 5. The PTM-wise prediction confidence range during large scale extraction is shown in Fig. 4, and confidence

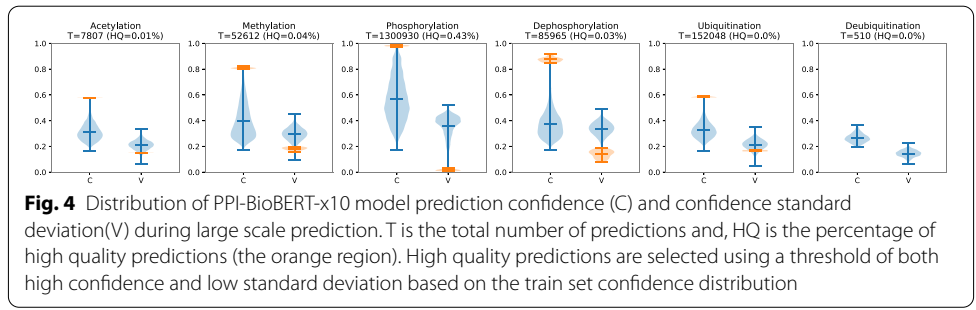

**Fig. 4** Distribution of PPI-BioBERT-x10 model prediction confidence (C) and confidence standard deviation(V) during large scale prediction. T is the total number of predictions and, HQ is the percentage of high quality predictions (the orange region). High quality predictions are selected using a threshold of both high confidence and low standard deviation based on the train set confidence distribution

**Table 5** The results of large scale prediction

| PTM | All | All (U) | HQ | HQ (U) | HQ MA | HQ MA (U) |
| --- | --- | --- | --- | --- | --- | --- |
| Acetylation | 7807 | 6113 | 1 | 1 | 0 | 0 |
| Dephosphorylation | 85965 | 50004 | 29 | 29 | 1 | 1 |
| Deubiquitination | 510 | 460 | 0 | 0 | 0 | 0 |
| Methylation | 52612 | 29914 | 20 | 18 | 4 | 2 |
| Phosphorylation | 1300930 | 381157 | 5654 | 4532 | 1659 | 537 |
| Ubiquitination | 152048 | 78859 | 4 | 4 | 0 | 0 |
| Total | 1599872 | 546507 | 5708 | 4584 | 1664 | 540 |

**All** predictions indicate all the predictions from PPI-BioBERT-x10. **U**nique represents unique PTM-PPI triplet predictions. **HQ** represents high quality PTM-PPI after thresholding. **HQ MA** represents high quality PTM-PPI available in multiple abstracts range is similar to training confidence ranges, where PTMs with lower training samples have lower confidence ranges. After applying interaction-wise probability and confidence standard deviation thresholds, we retain 5708 PPIs across 5 interaction types. From the 5708 predictions (4584 unique PTM-PPI triplets), we randomly select 30 predictions per PTM type for human verification. Despite the high precision (100.0%; see Table 4) on the test set after applying thresholds we find that during the large scale prediction the precision on the randomly sampled subset (subset in Additional file 1: Table A14) drops to 33.7 as shown in Table 6. With the estimated precision of 33.7%, of the 5708 $\approx$ 1900 have the potential to be correct PTM-PPI triplets.

**Table 6** Human evaluation on randomly sampled subset (30 interactions per PTM, unless there are fewer predictions) selected after thresholding on average confidence and standard deviation

|  | **Acety.** | **Dephosph.** | **Methy.** | **Phosph.** | **Ubiquit.** | **Total** |
| --- | --- | --- | --- | --- | --- | --- |
| Correct | 0 | 11 | 11 | 6 | 0 | 28 |
| Incorrect—DNA methylation | 0 | 0 | 2 | 0 | 0 | 2 |
| Incorrect—NER | 0 | 2 | 1 | 3 | 0 | 6 |
| Incorrect—no trigger word | 0 | 1 | 0 | 2 | 4 | 7 |
| Incorrect—opposite type | 0 | 1 | 0 | 0 | 0 | 1 |
| Incorrect—relationship not described | 0 | 14 | 4 | 19 | 0 | 37 |
| Not—sure | 1 | 0 | 1 | 0 | 0 | 2 |
| Total | 1 | 29 | 19 | 30 | 4 | 83 |

We find that percentage of precision error per PTM does not seem to be correlated to the number of training samples. For instance, methylation only has 10 training samples but has the highest precision (57%, 11 out of 19), whereas phosphorylation has the highest number of training samples but only 20% (6 out of 30) are correct.

We analyse the incorrect predictions further and categories them into:

- **Incorrect—DNA methylation:** The abstract describes DNA methylation instead of protein methylation. This type of error mainly affects methylation. For instance from the input abstract PubMed 19386523, the prediction <Q01196 (RUNX1), Methylation , Q06455(ETO)> whereas the abstract describes *"RUNX1-ETO fusion gene on DNA methylation"*
- **Incorrect—NER:** NER has either not identified proteins mentions or not normalised them correctly.
- **Incorrect—no trigger word:** The abstract does not even mention the trigger word describing the predicted PTM. For instance,the prediction `<P48431 (Sox2), Ubiquitination, Q01860 (Oct3/4)>` where the input abstract PubMed 22732500 does not mention the trigger *ubiquit.*, but the abstract mainly describes *"knockdown of Sox2 and Oct3/4 gene expression in HCC cells can reduce carboplatin-mediated increases in sphere formation and increase cellular sensitivity to chemotherapy"*.
- **Incorrect—opposite PTM:** The abstract mentions the complementary PTM, e.g. (phosphorylation, dephosyphorylation), (ubiquitination, deubiquitinaion). For instance, the prediction `<Q15746 (myosin light chain kinase), dephosyphorylation, Q7Z406 (myosin)>` from the abstract PubMed 2967285, the abstract describes phosphorylation - *"phosphorylation of the dephosphorylated brain myosin with myosin light chain kinase and casein kinase II"*
- **Incorrect - Relationship not described:** This is effectively any other type of error that doesn't fall into any of the above categories. This category has the largest percentage of error (70%, 37 out of 53 incorrect predictions). An example prediction in this category is <P0870 (IL3), Phosphorylation, P36888 (FLT3)> from the PubMed abstract 10720129 doesn't describe phosphorylation PTM between IL3 and FLT3 but rather *"Somatic mutation of the FLT3 gene, in which the juxtamembrane domain has an internal tandem duplication, is found in 20% of human acute myeloid leukemias and causes constitutive tyrosine phosphorylation of the products. In this study, we observed that the transfection of mutant FLT3 gene into an IL3-dependent murine cell line, 32D, abrogated the IL3-dependency."*
- **Incorrect—not related to PPI:** Here prediction is from an abstract that is not even related to PPI. This is the worst form of false positive error. An example of this is a prediction `<P16070 (CD44), phosphorylation, P60568(IL2)>` from PubMed abstract 7763733.
- **Unsure:** In this scenario, the human reviewer is unable to decide if the abstract describes the PTM-PPI by solely reading the abstract. An example of this is a predic-

**Table 7** Include multiple abstracts filter: human evaluation on randomly sampled subset selected after thresholding on average confidence and standard deviation, with the additional condition that these predictions are present in multiple abstracts

|  | **Methyl.** | **Phosphoryl.** | **Total** |
| --- | --- | --- | --- |
| Correct | 4 | 16 | 20 |
| Incorrect—NER | 0 | 2 | 2 |
| Incorrect—Not related to PPI | 0 | 1 | 1 |
| Incorrect—relationship not described | 0 | 7 | 7 |
| Not—sure | 0 | 4 | 4 |
| Total | 4 | 30 | 34 |

We select 30 interactions per PTM, unless there are fewer predictions

**Table 8** Comparison of PPI-BioBERT-x10 predictions with iPTMnet

| **PTM** | **iP Total** | **iP Unique** | **iP Uniprots** | **Ours** | **Ours HQ** | **iP RLIMS** |
| --- | --- | --- | --- | --- | --- | --- |
| Acetylation | 141 | 73 | 12 | 0 | 0 | 0 |
| Methylation | 7 | 4 | 4 | 0 | 0 | 0 |
| Phosphorylation | 21050 | 8949 | 8805 | 3270 | 815 | 358 |
| Ubiquitination | 2 | 1 | 0 | 0 | 0 | 0 |

Of all PTM-PPI entries in iPTMnet *(iP Total)*, *iP Unique* represents the subset of unique entries. Of the unique PTM-PPIs the subset that has associated UniProt identifiers is in column *iP Uniprots*. *iP RLIMS* is the number of unique PPI-PTM sourced from RLIMS+. The number of all the PPI-BioBERT-x10 predictions that can be recalled in iPTMnet is in *Ours*. *Ours HQ* represents the High Quality PPI-BioBERT-x10 predictions after confidence thresholding

```
tion <P01019 (Angiotensin II), phosphorylation, P30556(AT1)>
```
from PubMed abstract 9347311.

In order to reduce the false positive predictions, we further filter the high confidence and low variation predictions to a subset of predictions that are available in at least 2 abstracts to select 1659 predictions (537 unique PTM-PPI) and verify a randomly selected subset (30 per PTM) of predictions. We find that the precision on a randomly sampled subset (available as Additional file 1: Table 13) now improves to 58.8% as shown in Table 7. The intuition behind this improvement is if the same PTM-PPI prediction can be inferred from multiple papers, the probability of the prediction being right is higher.

We also compare the recall of the predictions with those PTM-PPI consolidated in iPTMnet [20][1] as shown in Table 8. The iPTMnet paper [20] mentions that the iPTMnet database is updated on a monthly basis using text mining, leveraging RLIMS-P [17] and eFIP [16], processing all the abstracts and full length articles available in PubMed Central Open Access. Hence, we presume that our results, obtained using the entire set of PubMed abstracts available at the time of our study, are comparable to the text-mined PTM-PPIs in iPTMnet. Using PPI-BioBERT-x10, we have identified 37.1% (3270 out of 8805) of iPTMnet PTM-PPIs, with 815 of these found in the high confidence region. Only a total of 358 PTM-PPI in iPTMnet were sourced directly from literature using

---

[1] https://research.bioinformatics.udel.edu/iptmnet_data/files/current/ptm.txt.

**Table 9** Results of noise levels, after noise reduction, in training data to verify if the PPI relationship is described in the abstract

| Interaction type | Correct | Not—sure | Total |
| --- | --- | --- | --- |
| Acetylation | 4 | 1 | 5 |
| Dephosphorylation | 6 | 4 | 10 |
| Deubiquitination | 1 | 1 | 2 |
| Methylation | 4 | 6 | 10 |
| Phosphorylation | 6 | 4 | 10 |
| Ubiquitination | 2 | 3 | 5 |
| Total | 23 | 19 | 42 |

The training data is randomly sampled (10 samples per interaction type unless the number of available training samples is lower) and verified by a human annotator text mining, specifically the rule-based text mining method RLIMS+ [17], compared to the 3270 (815 after confidence thresholding) identified by PPI-BioBERT-x10. Hence, the recall of PPI-BioBERT-x10 in relation to the PTMs captured in iPTMnet is substantially higher than the RLIMS+ method, by over 200% after confidence thresholding, demonstrating the robustness of our machine learning-based approach.

## Discussion

### Noise in distant supervision

There are two types of noise in our distantly supervised dataset: **(a)** false positive noise and **(b)** false negative noise. False positive noise occurs where a given PPI relationship <Protein1, *PTM function*, Protein2> may not be described in the abstract but is labelled as describing the relationship. False negative noise occurs where the relationship <Protein1, *PTM function*, Protein2> is in fact described in the abstract but is not labelled as describing the relationship.

Our simple heuristic removes noisy training samples, more specifically false positive noise, i.e., if the normalised Uniprot identifiers of the participating entities are not mentioned in the abstract and the abstract is not likely to describe the relationship (unless NER has failed to detect the protein names and normalise them). We manually inspected the quality of training data after noise reduction using a randomly sampled subset, as detailed in Table 9. While the noise reduction heuristic has been fairly effective in removing false positive noise, in 45% of the cases the human reviewer (author A.E. with basic knowledge on PPIs and PTMs) was unable to decide whether the PPI relationship is described or not in the abstract. An example of a sample marked as unsure, is phosphorylation between c-Jun N-terminal kinase (JNK) and c-Jun (IntAct entry https://www.ebi.ac.uk/intact/interaction/EBI-7057279) in the abstract PubMed 19527717. This shows the complexity of the task beyond language, the need for prior context or assumed domain knowledge on how PPIs interact to interpret the abstracts for manual annotation. In the test set, we found that a combination of high confidence and low variation in confidence has been able detect some of the incomplete annotation (false negatives) as shown in Table 4.

One of the main challenges is that it is difficult to quantify noise without manual verification. Hence, machine learning approaches that rely on distant supervised data need to manually verify label quality, at the very least, on randomly subset of the data. This requires a user interface (UI) that makes it easy for humans to review the samples effectively. For this work we built a UI (see Fig. 5) using Amazon SageMaker Ground Truth, to assist in verification of both the labelled and the large scale prediction samples.

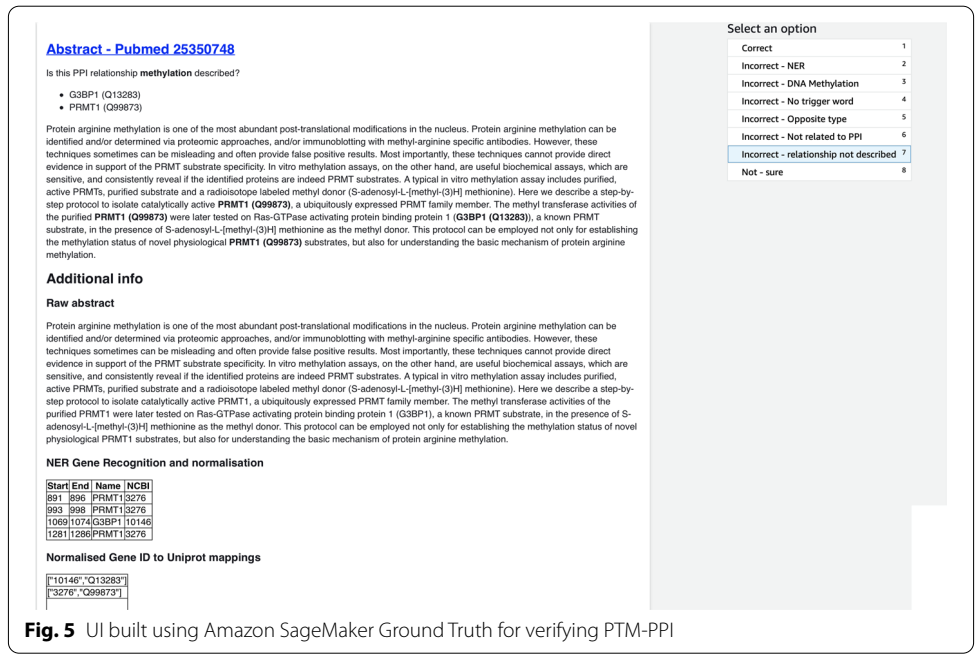

**Fig. 5** UI built using Amazon SageMaker Ground Truth for verifying PTM-PPI

**Generalisability of PTM-PPI extraction**

Generalisability in machine learning is a broadly discussed issue. Futoma et al. [24] in their work in a clinical context describe how important it is to understand how, when, and why a machine learning system works. In this paper, we primarily investigated the use of confidence calibration as an indicator of when a prediction is valid. We hypothesised that using the average ensemble confidence, a state-of-the-art uncertainty estimation approach by Lakshminarayanan et al. [6], with our extension to also include low variation to counteract the effects of class imbalance (where each PTM has high difference in the number of training samples) would minimise incorrect predictions. While this hypothesis seems to hold true in the test and validation tests, the results of our large-scale application experiment, as shown in Table 6, demonstrate that the model makes many errors in that scenario.

To understand the drop in precision, we compare the cosine similarity (where the abstract is represented as a count vector of unigrams) of each test sample abstract with the closest match in the train set using the approach followed by Elangovan et al. [23]. We also compare the similarity of the abstracts from the large scale predictions with the train and test set. In order to study the effects of confidence calibration, we divide the large scale predictions into 2 groups (a) the $\approx 5700$ high quality predictions that have high confidence score and low standard deviation as detailed in section *Large scale PTM-PPI extraction from PubMed abstracts* and in Table 5 (b) the predictions that have low confidence scores (PTM-wise threshold less than minimum training confidence) and high standard deviation (PTM-wise threshold greater than maximum training standard deviation), which we refer to as low quality predictions. We find that the high quality predictions have abstracts that are quite similar to the test set and to the train set as shown in Fig. 6. This is in contrast to the low quality predictions, whose abstracts are less similar to the train and test set as shown in Fig. 6. Furthermore, in the high quality predictions the common words for each interaction type are "key terms" for that interaction type (e.g. the abstracts associated with phoshorylation predictions have top common words such as kinase and phosphorylation) as shown in Fig. 7, whereas the low quality predictions have generic words such as expression and activity as the most common words. This shows that the model has higher confidence on abstracts that are more relevant for that interaction type.

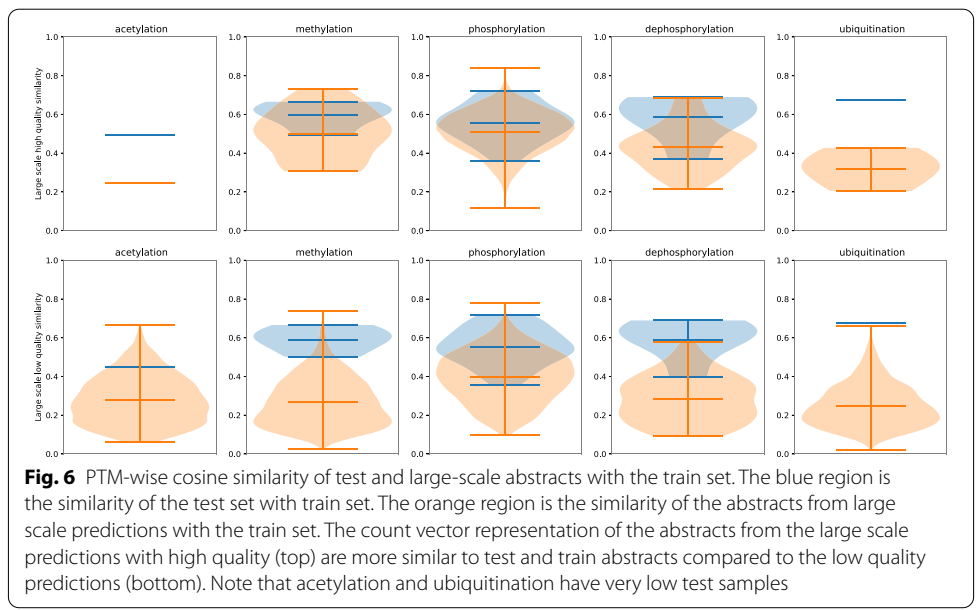

**Fig. 6** PTM-wise cosine similarity of test and large-scale abstracts with the train set. The blue region is the similarity of the test set with train set. The orange region is the similarity of the abstracts from large scale predictions with the train set. The count vector representation of the abstracts from the large scale predictions with high quality (top) are more similar to test and train abstracts compared to the low quality predictions (bottom). Note that acetylation and ubiquitination have very low test samples

Although the model seems to be fairly successful in detecting the right abstracts, it seems to struggle to classify the correct relationship between protein pairs and the PTM, as seen in the manual verification results in Table 6. This indicates that the model has learnt shallow features and not the semantic relationships between the proteins represented using markers (`PROTPART1, PROTPART2`). We also find that even low quality predictions for acetylation seem to be derived from abstracts containing relevant words, based on the analysis in Fig. 7, which may suggest that human curators can relax the quality thresholds to add more acetylation PPIs to the knowledge-base.

Previous papers on PPI extraction using machine learning [13, 26, 32] have evaluated the efficacy of their method on a test set, and have used the performance on the the test set as an indicator of generalisability. The method used to create a test set and its impact on assessing generalisability has been studied in previous papers [23, 33]. For instance, our prior work [23] has identified data leakage using random splits with 10-fold cross validation on the PPI interaction AIMed dataset where the random split results in over 70% tri-gram word overlap between train and test folds resulting in inflated performance on the test set.

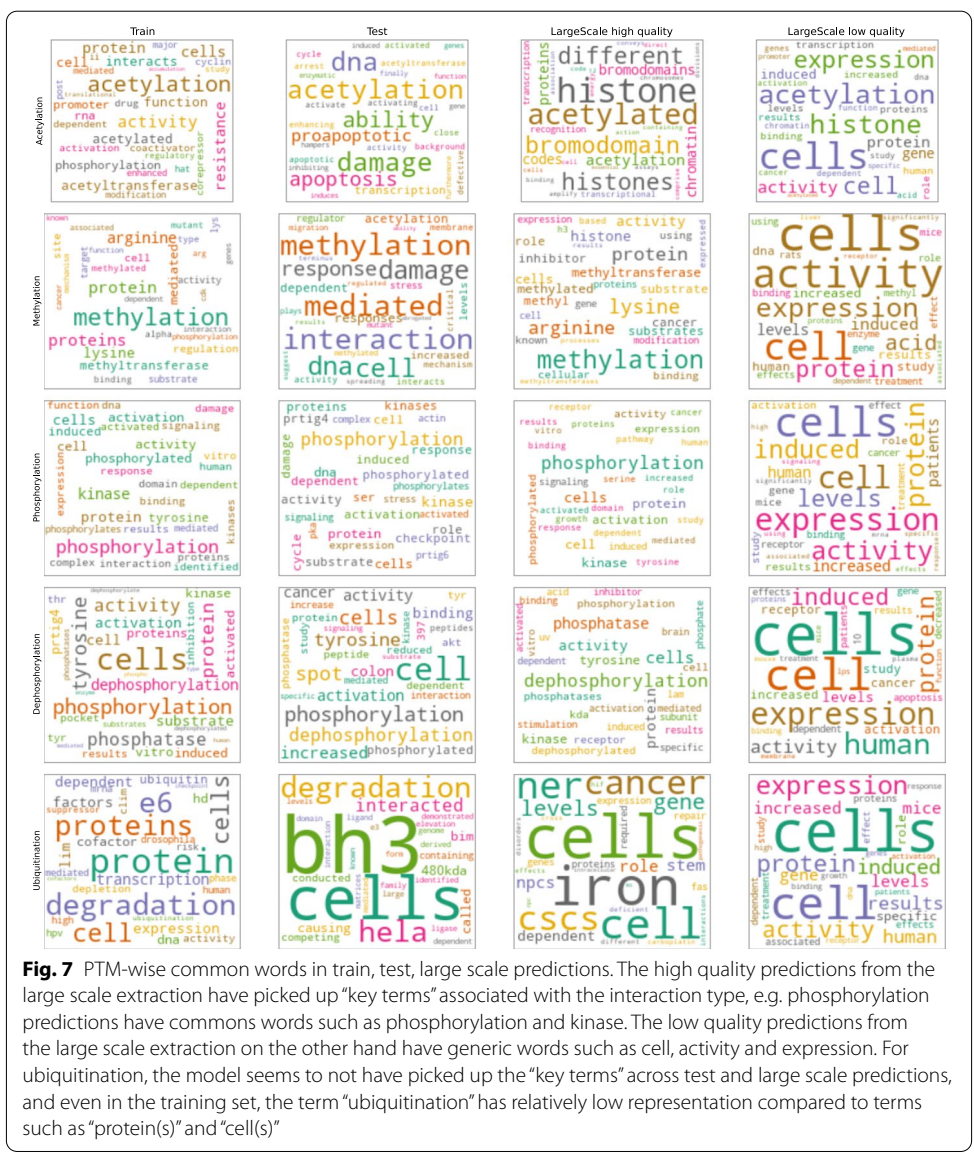

**Fig. 7** PTM-wise common words in train, test, large scale predictions. The high quality predictions from the large scale extraction have picked up "key terms" associated with the interaction type, e.g. phosphorylation predictions have commons words such as phosphorylation and kinase. The low quality predictions from the large scale extraction on the other hand have generic words such as cell, activity and expression. For ubiquitination, the model seems to not have picked up the "key terms" across test and large scale predictions, and even in the training set, the term "ubiquitination" has relatively low representation compared to terms such as "protein(s)" and "cell(s)"

Substantially lower real world performance compared to a test set, specifically in the context of BERT models, has also been reported in previous work including McCoy et al. [33]. In that work, they experiment with an entailment task, fine-tuning 100 BERT models which only vary in (a) the random initial weights of the classifier and (b) the order in which training examples were presented to the model. They find that even simple cases of subject-object swap (e.g., "determining that the doctor visited the lawyer" vs "the lawyer visited the doctor"), resulted in accuracy variation between 0.0 and 66.2% across the BERT models, whereas the test-set performance remained fairly consistent between 83.6% and 84.8%. In addition, previous studies in computer vision have also shown neural networks can provide high confidence prediction even when the model is wrong despite robust performance on the test set [7, 34]. In particular a type of model uncertainty, epistemic uncertainty [34], where the model doesn't have sufficient information to make a decision yet has made overconfident predictions continues to be a challenge despite applying machine learning methods. In the context of our work, the need for diverse and effective negative samples is pertinent, given the number of false positive predictions despite relatively high confidence. The creation of effective negative samples and robust generalisability remain open research questions in need of further study.

### Human curation augmentation for PTM-PPI extraction

The prediction quality can be improved by providing more training samples [34, 35], which requires manual curation. However, manual curation is difficult, time consuming and not cost effective [3, 4]. This becomes a chicken and egg problem: automation is meant to speed up curation and reduce manual curation effort, however we need large amounts of manually curated training data so that the models can produce sufficiently reliable predictions.

Triaging journals alone takes up approximately 15% of the curation time [36] to select relevant papers, while the rest of the tasks such as named entity recognition and normalisation, detecting relationships take up the rest 85%. In our research, we surface PTM-PPI triplets where the entities are normalised. Hence, our solution supports human augmentation [37], i.e.using machine learning to make it much easier for humans to curate. This is one of the major advantages of machine learning compared to rule based systems, as adding more data and retraining can improve the model without manually rewriting the rules. The second advantage of using confidence thresholding is the ability to adjust the thresholds per PTM enabling faster curation of PTM-PPIs with reduced representation, such as for relatively rare interaction types like deubiquitination. In order to take advantage of machine learning, we need to ensure human annotators have: **(a)** a convenient user interface so the curator can quickly accept or reject a prediction and **(b)** provide a high hit ratio of correct PPIs using confidence thresholding so that humans can add more PPIs to the knowledge bases faster. This will enable continuous retraining the model with more diverse training data (both positive and negative samples) so the model can evolve and further reduce manual curation effort over time.

### Conclusion

We created a distant supervised training dataset for extracting PTM-PPIs, including annotation for phosphorylation, dephosphorylation, methylation, demethylation, ubiquitination, and acetylation from PubMed abstracts by leveraging the IntAct database.

Using this dataset we trained an ensemble model, PPI-BioBERT-x10, and applied an approach by Lakshminarayanan et al. [6] to improve confidence calibration of the neural network by averaging the confidence scores predicted by the ensemble. We find that the predicted confidence range for each PTM is proportional to the number of training samples available and the predicted confidence range is consistent between train, test validation and large scale predictions. As a result, we extended the work by Lakshminarayanan et al. [6] using a threshold per PTM by combining average confidence with confidence standard deviation to improve confidence calibration and counteract the effects of class imbalance during calibration, resulting in 100% precision (retaining 19% of the positive predictions) in the test set.

We applied PPI-BioBERT-x10 to text-mine 18 million PubMed abstracts, extracting 1.6 million PTM-PPI predictions ($\approx$ 540,000 unique PTM-PPI ) and shortlisting approximately 5700 ($\approx$ 4500 unique PTM-PPI ) PTM-PPIs with relative high confidence and low variation. However, on manual review of a randomly sampled subset we find that the precision of prediction drops to 33.7% and generalisability aspect of knowing when a prediction is correct remains a challenge. The main source of precision error seems to lie in associating the correct the protein pairs and the PTM, while the abstract itself seems fairly relevant to PTM-PPI extraction indicating that the model has learnt shallow features and not the semantic relationships necessary for relation extraction. By selecting predictions that appear in multiple abstracts ($\approx$ 1600 triplets), the precision on a randomly sampled subset improves to 58.8%. We also find that PPI-BioBERT-x10 is able to identify over 200% more PTM-PPIs compared to RLIMS+ [17] a rule based system, revealing the advantage of PPI-BioBERT-x10 with regards to recall compared to a manually crafted rule based system. We also propose that to decrease manual curation effort and to improve the prediction quality over time requires using an effective user interface along with confidence thresholding to allow humans curators to easily accept or reject predictions and continuously retrain the model.

In this work, we studied the advantages and challenges of adopting deep learning-based text mining for a new task, PTM-PPI triplet extraction, using distantly supervised data. This work highlights the need for effective confidence calibration, the importance of generalisability beyond test set performance and how real world performance can vary substantially compared to the test set. Further study includes the use of effective adversarial samples to improve the robustness of machine learning models for practical use.

#### Abbreviations
NER: Named entity recognition; PPI: Protein protein interaction; PTM: Post translational modification; ECE: Expected calibration error; NLP: Natural language processing; BERT: Bidirectional encoder representations from transformers.

## Supplementary Information
The online version contains supplementary material available at https://doi.org/10.1186/s12859-021-04504-x.

**Additional file 1.** Appendix data and explanations.

#### Authors' contributions
AE carried out computational experiments and was responsible for the design, data analysis, data interpretation, developing software, discussion and drafting and editing the manuscript. AE, MD, KV conceptualised the study. MD, YL, DP and KV supervised AE, were involved in the design and execution of the study, supported data interpretation, and contributed to writing the manuscript. All authors read and approved the final manuscript.

#### Funding
This work is supported by Australian Research Council Grants DP190101350 (KV) and LP160101469 (KV,YL).

#### Availability of data and materials
We publish our source code, training, test and validation dataset and the predicted $\approx 5700$ PPIs on https://github.com/elangovana/large-scale-ptm-ppi. The Amazon SageMaker Ground Truth UI code is available at https://github.com/elangovana/ppi-sagemaker-groundtruth-verification.

## Declarations

**Ethics approval and consent to participate**
Not applicable.

**Consent for publication**
Not applicable.

**Competing interests**
We certify that there are no financial and non-financial competing interests regarding the materials discussed in the manuscript.

**Author details**
[1]School of Computing and Information Systems, The University of Melbourne, Melbourne, Australia. [2]The Walter and Eliza Hall Institute of Medical Research, Melbourne, Australia. [3]Department of Clinical Pathology, Faculty of Medicine, Dentistry and Health Sciences, The University of Melbourne, Melbourne, Australia. [4]School of Computing Technologies, RMIT University, Melbourne, Australia.

## Publisher's Note

Springer Nature remains neutral with regard to jurisdictional claims in published maps and institutional affiliations.